\documentclass[journal]{IEEEtran}
\usepackage{cite}

\ifCLASSINFOpdf
  \usepackage[pdftex]{graphicx}
\else
  \usepackage[dvips]{graphicx}
\fi
\usepackage{amsmath}
\usepackage{algorithmic}
\usepackage{url}

\usepackage{booktabs}
\usepackage{multirow}
\usepackage {diagbox}
\usepackage {colortbl,xcolor}
\usepackage[colorlinks,
            linkcolor=blue,
            anchorcolor=blue,
            citecolor=blue]{hyperref}
\usepackage{algorithm}

\begin{document}
%
\title{Joint Low-level and High-level Textual Representation Learning with Multiple Masking Strategies}
%
%
%


\author{Zhengmi~Tang,~\IEEEmembership{Member,~IEEE,}
        Yuto~Mitsui,
        Tomo~Miyazaki,~\IEEEmembership{Member,~IEEE,} \\
        and~Shinichiro~Omachi,~\IEEEmembership{Senior Member,~IEEE}
\thanks{This work was partially supported by JSPS KAKENHI under Grants JP22H00540 and JP22K12729 and "Pioneer" and "Leading Goose" R\&D Program of Zhejiang (2025C01222). \it{(Zhengmi Tang and Yuto Mitsui contributed equally to this work.)}}
\thanks{Zhengmi Tang is with Wenzhou University Artificial Intelligence and Advanced Manufacturing Institute (AIAMI), Wenzhou City, China (e-mail: tzm@dc.tohoku.ac.jp).}
\thanks{Yuto Mitsui, Tomo Miyazaki, and Shinichiro Omachi are with the Graduate School of Engineering, Tohoku University, Sendai, Japan (e-mail: yuto.mitsui.s1@dc.tohoku.ac.jp; tomo@tohoku.ac.jp; machi@ecei.tohoku.ac.jp).}

}

\maketitle

\begin{abstract}

Most existing text recognition methods rely on large-scale synthetic datasets for training due to the scarcity of labeled real-world datasets. However, synthetic data falls short in naturally replicating various real-world scenarios, such as uneven illumination, irregular layout, occlusion, and image degradation, resulting in performance disparities when handling complex real-world text images. To tackle this issue, self-supervised learning techniques, such as contrastive learning, and mask image modeling (MIM), have emerged to leverage unlabeled real images to bridge the domain gap in text recognition tasks. This study investigates the textual representation in the original Masked AutoEncoder (MAE) and reveals that MAE with random masking predominantly captures low-level textural features, lacking efficiency in extracting high-level semantic representations from text images. To fully exploit the potential of masked image modeling for text recognition, we delve into the contextual information inherent in text images by introducing random blockwise masking and span masking. Unlike random patch masking, which discretely masks image patches, blockwise masking and span masking enable the continuous masking of image patches, leading to the complete removal of some characters. These approaches compel the model to explicitly learn the contextual relationships between characters in a word image. By integrating random patch masking, blockwise masking, and span masking for MIM, our Multi-Masking Strategy (MMS) facilitates the joint learning of both low and high-level representations, enhancing the effectiveness of textual representation learning. The comprehensive experimental results demonstrated that MMS outperforms the state-of-the-art self-supervised methods in various text-related tasks, including text recognition, text segmentation, and text image super-resolution when fine-tuned with real data.

\end{abstract}

\begin{IEEEkeywords}
Scene text recognition, self-supervised learning, masked image modeling.
\end{IEEEkeywords}

%
\IEEEpeerreviewmaketitle

\section{Introduction}

\IEEEPARstart{S}cene Text Recognition (STR) is a crucial task that focuses on reading text in natural scenes and finds a wide range of applications, such as navigation in automated driving \cite{AutomaticDA2004}, translation of signs and menus, content-based image retrieval \cite{zhu2017cascaded}, etc.
While the field of optical character recognition (OCR) has made significant advancements with the assistance of deep learning, STR remains a challenging task due to diverse fonts, text shapes, and the environmental conditions in image capture.
Most existing text recognition methods are trained using large synthetic datasets \cite{mj,gupta_synthetic_2016,LongUnreal2020,TangLBTS2022}, primarily due to the limited availability of labeled real-world datasets. 
However, these methods struggle to address real-world problems due to the domain gap between synthetic and real data.
Hence, there is a growing interest in utilizing self-supervised learning methods to pre-train text recognition models by leveraging unannotated real images.

Contrastive learning and masked image modeling have been introduced as self-supervised learning methods. 
Contrastive learning leverages discriminative pretext tasks, such as applying data augmentations on different views, to extract latent features that are invariant to the augmentations. Consequently, the data augmentation pipeline plays a significant role in current contrastive learning and is mostly based on aggressive cropping, flipping, color distortions, and blurring. However, unlike object images used in object classification, where the entire image represents a single class (semantic) property, a text image consists of a sequence of characters, and the atomic elements of text images should be characters. In the context of sequence-level text representation learning methods, directly applying strong geometric transformations from conventional schemes may result in character misalignment issues between different views. To this end, SeqCLR \cite{seqclr} models text images as a sequence of adjacent image slices and horizontally splits the feature to obtain multiple comparison elements for contrast learning. It also utilized constrained data augmentations to preserve the sequence information. 
PerSec \cite{persec} introduces hierarchical contrastive learning on low-level stroke and high-level semantic contextual spaces to explore the visual and semantic properties contained in text images.
CCD \cite{ccd} proposes a feature-level character alignment strategy to achieve character-level contrast elements for contrastive learning. This approach utilizes the augmentation matrix of color images and character masks. Character masks are generated by a self-supervised character segmentation module, which extracts character structures from unlabeled real images using pseudo labels from K-means.
DiG \cite{dig} integrates contrastive learning and masked image modeling into a unified model. It applies random patch masking to one view of contrastive learning, thus taking advantage of both discrimination and generation for text recognition. Yet the data augmentation pipeline employed in DiG follows that of SeqCLR \cite{seqclr}.

\begin{figure}[t]
\centering
\includegraphics[]{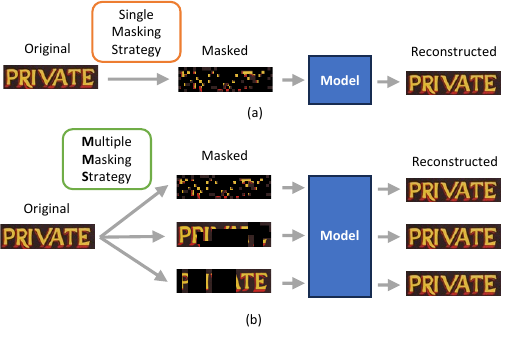}
\caption{Illustration of (a) existing mask image modeling methods and (b) our proposed MMS that can use multiple masking strategies.}
\label{fig:comparison mae}
\end{figure}

Masked image modeling (MIM) does not require aggressive data augmentation. However, masking strategy, masking ratios, and patch sizes are critical for MAE to learn succinct and comprehensive object information. In the context of object classification tasks, Kong \textit{et al.} \cite{understanding-mae} found that random patch masking in MAE \cite{mae} faces challenges in learning high-level representations and often yields relatively low-level representations. These low-level representations mainly capture texture information, which can be predicted using surrounding visible pixels, like interpolation. On the contrary, high-level representations encompass semantic information, and they cannot be effectively captured without understanding the meaning of the image. 
Text images consist of character sequences, where textural (stroke), and semantic (character) information are contained. Stroke information is low-level information that explicitly differentiates the text foreground from the background, while character information pertains to high-level information that allows for the identification of individual character instances. 
Considering these characteristics, we assert that random patch masking is not efficient for extracting high-level representation and fails to fully exploit the potential of masked image modeling for text recognition.

In this study, we investigate the mining of high-level representation for text recognition by considering the unique contextual information present in text images. Characters are the atomic elements with individual semantic meanings, but when they form word images, contextual (linguistic) information is embedded within the images. 
To utilize the contextual information, we investigate random blockwise masking and span masking in the framework of MAE. Blockwise masking generates a mask consisting of several rectangle blocks with random block size and aspect ratios and span masking generates a mask with multiple horizontal widths. Different from random patch masking, which masks the image patches discretely, blockwise masking and span masking can mask continuous patches, leading to the removal of a complete or substantial portion of some characters, thereby forcing the network to explicitly learn contextual information between the characters in a word image.
Furthermore, we integrate random patch masking, blockwise masking, and span masking as \textbf{M}ulti-\textbf{M}asking \textbf{S}trategy (MMS) for MIM to facilitate efficient and joint learning of both low and high-level textual representations. Fig~\ref{fig:comparison mae}(b) shows our concept.


The main contributions of this paper are as follows.
\begin{enumerate}
\item We investigate different masking strategies for mask image modeling in self-supervised textual representation learning, and find random patch masking predominantly captures low-level textural features, while blockwise masking and span masking can model high-level semantic representations.
\item We propose a simple yet efficient Multi-Masking Strategy (MMS) for text recognition, which combines random patch masking, block masking, and span masking to jointly learn low-level textural and high-level semantic representations from text images.
\item The experimental results demonstrate the significant superiority of MMS in self-supervised representation learning for various text-related tasks, including text recognition, text segmentation, and text image super-resolution. Models pre-trained with MMS outperform the state-of-the-art self-supervised methods when fine-tuned with real data.
\end{enumerate}


\begin{figure*}[t]
\centering
\includegraphics[width=0.95\textwidth]{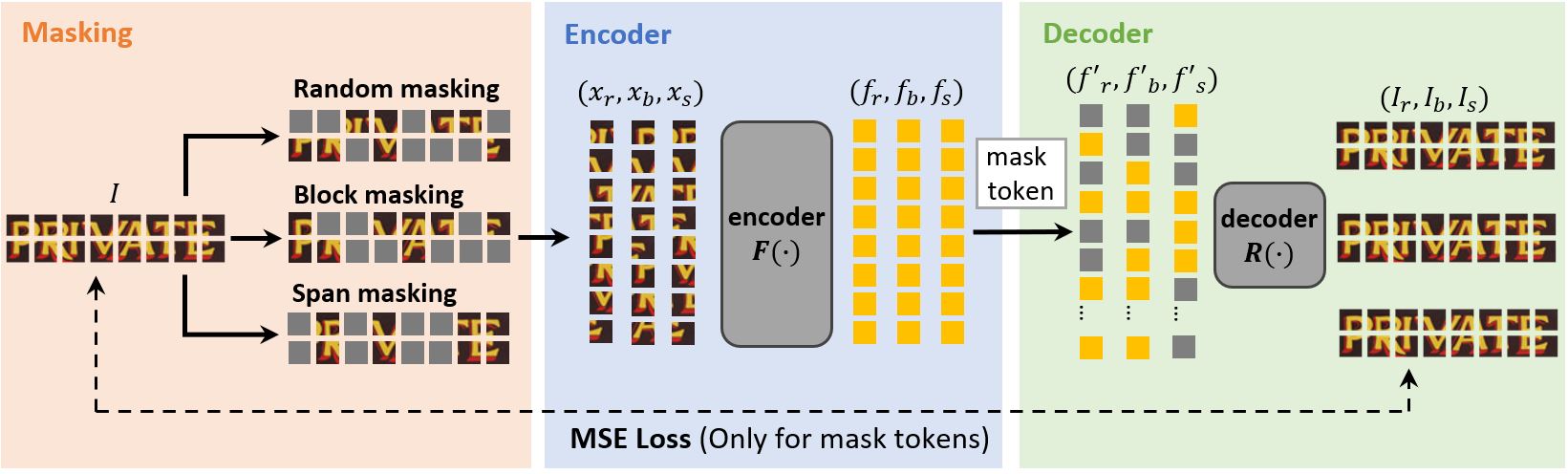}
\caption{A framework of Multi-Masking Strategy (MMS). Encoder and decoder parameters are shared between branches. During pre-training, a subset of image patches is masked (removed) by random patch masking, block masking, and span masking, respectively. The encoder only processes a subset of the visible patches. The decoder reconstructs images from the encoder output and mask tokens.}
\label{fig:arch}
\end{figure*}

\section{Related Work}
\label{Related Work}
\subsection{Text Recognition}
Scene text recognition (STR) predicts the character sequence in a text image, typically a text-centered image cropped from a text region within a scene text image.
In the deep learning era, STR models are commonly categorized into context-free (visual) and context-aware (language) methods.

Context-free studies focus on visual information and directly predict the characters based on image features, with the output characters independent of each other. 
Rectification-based methods \cite{Shi_2016_RARE,aster,zheng2023tps++} utilize differentiable Thin-Plate-Spline (TPS) transformation \cite{JaderbergSTN2015} to rectify irregular text images to regular ones, facilitating feature extraction for the recognition model.
Segmentation-based methods \cite{lyu2018masktextspotter,liao2019scene,wan2020textscanner} leverage character-level bounding box annotation to segment character regions and subsequently predict the final character sequence.
Additionally, some studies introduce implicit attention mechanisms in either 1D \cite{FocusingAttention,liu2018squeezedtext,qin2021dynamic} or 2D space \cite{li2019show,dai2020sloan,fang2021read,mgp-str,sgbanet,siga,satrn,corner-transformer} to obtain spatial character features by computing the similarity between feature patches.
For example, 
SGBANet\cite{sgbanet} initially uses semantic GANs to produce basic semantic features and then employs a balanced attention module to perform recognition. 
SIGA \cite{siga} improves attention accuracy in recognition by delineating glyph structures of text images through self-supervised text segmentation and implicit attention alignment.
SATRN\cite{satrn} proposes 2D self-attention in the Transformer to capture long-range 2D spatial dependencies among characters in scene text images.
CornerTransformer\cite{corner-transformer} employs corner points for recognizing complex artistic text and models the global relationship between image features and corner points through cross-attention.
 
Context-aware methods utilize language models to incorporate text semantic information for refining predictions.
ABINet \cite{abinet} restricts gradients when passing the visual model's output to the language model, ensuring independence between the visual and language models and enabling iterative text modification.
LevOCR \cite{levocr} employs LevT \cite{levenshtein-transformer}, a technique from the field of language processing that explicitly handles token addition and deletion, in its language model.
PARSeq \cite{parseq} trains language models using PLM (Permutation Language Modeling), demonstrating fast inference speed and competitive performance compared to existing methods.
\subsection{Visual Self-supervised Learning}
In recent years, self-supervised learning has achieved great success in computer vision. Self-supervised learning methods can learn image representations on pretext tasks, whereas the representative techniques are discriminative tasks of contrastive learning (CL) and generative tasks of masked image modeling (MIM).

Contrastive learning methods learn the visual representation by discriminating image similarity between positive and negative views, generated through data augmentations.
MoCo \cite{moco} uses a large queue to store negative samples so that it can take in more negative examples for CL. MoCo also introduces a momentum encoder to ensure the consistency of the samples in the queue.
SimCLR \cite{simclr} simplified CL framework by removing specialized architectures or a memory bank and emphasized the data augmentations and larger batch size.
BYOL \cite{byol} learns its representation by predicting previous versions of its outputs, without using negative pairs.
SimSiam \cite{simsiam} successfully replaces the momentum updating technique with a stop-gradient technique.
MoCo-v3 \cite{moco-v3} and DINO \cite{dino} extend previous methods and use vision transformers as the backbone for CL.

Masked image modeling, inspired by BERT \cite{DevlinBERT2019} in NLP, learns image representation by recovering the image from the partially masked image, where the learning target can be pixel-level or features/tokens-level.
BEiT \cite{beit} applies random block masking on some proportion of image patches and predicts the visual tokens of the original image obtained by a pre-trained discrete VAE \cite{DVAE2021}. 
MAE \cite{mae} first masks random patches of the image with a high masking ratio (75\%). Then, it only feeds visible image patches into an encoder and reconstructs the image pixels from the latent representations of the encoder and mask tokens with an auxiliary decoder. 
SimMIM \cite{simmim} takes mask tokens and image patches as the input of an encoder and reconstructs the raw pixels of the image with a lightweight linear head.
MaskFeat \cite{maskfeat} changes the prediction target of SimMIM to some hand-crafted features and reveals HOG descriptor is an effective target for MIM.
MAGE \cite{LiMAGE2023} brings the MAE framework with variable masking ratios into the latent token space modeled by VQVAE \cite{VQVAE2017} to unify the generative model and representation learning.

\subsection{Self-supervised Learning for Text Recognition}
Self-supervised learning pipelines have gained considerable attention in learning textual representation using scene text images without labels due to their promising results in OCR-related downstream tasks, such as text recognition, text segmentation, and text image super-resolution. 

SeqCLR \cite{seqclr} first expands the CL framework to visual sequence-to-sequence predictions in text recognition by dividing feature maps into a sequence of individual elements for contrastive loss. 
PerSec \cite{persec} proposes hierarchical contrastive learning, which can simultaneously contrast and learn latent representations from low-level stroke and high-level semantic contextual spaces.
DiG \cite{dig} proposes a method that combines CL and MIM. This method masks one of the views of CL and reconstructs the image with the pipeline of SimMIM \cite{simmim}.
CCD \cite{ccd} introduces a feature-level character alignment strategy to achieve character-level contrast elements for CL. This solves the problem of existing methods of inconsistent character-level features and inflexible data augmentation by creating a sequence of feature vectors horizontally from text images.
MaskOCR \cite{maskocr} explores a unified vision-language pre-training for the encoder-decoder recognition framework. It pre-trains the encoder using a large set of unlabeled text images to learn strong visual representations and directly pre-trains the sequence decoder to improve language modeling capabilities.

In terms of MIM-based self-supervised learning for text recognition, our method is closely related to the DiG and MaskOCR. DiG exploits random patch masking for MIM to assist CL while MaskOCR only utilizes random span masking for representation learning. They only take into account one aspect among low-level visual and high-level semantic representations for text recognition. 
In this study, we first analyze the characteristics of current MIM methods and investigate various masking strategies to explore the unique contextual features of text images. We integrate those masking strategies into one framework and propose a simple yet efficient learning method MMS. MMS combines random patch masking, span masking, and block masking, which can jointly learn both low and high-level textual representations. 

\section{Methodology}
\label{Methodology}
In this section, we introduce the Multiple Masking Strategy (MMS) for self-supervised textual representation learning. Following the masking-reconstruction paradigm \cite{mae} and the general pipeline of self-supervised learning \cite{ssl-survey}, our model comprises an encoder for extracting latent representations and task-specific decoders for various tasks, such as text image reconstruction, text recognition, text segmentation, and text super-resolution. 
Initially, the encoder and image reconstruction decoder are pre-trained on unlabeled datasets. Subsequently, for each downstream task, the pre-trained encoder and corresponding task-specific decoder are fine-tuned using the respective labeled data.




\begin{figure}[t]
\centering
\includegraphics[]{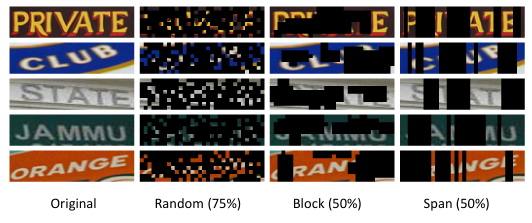}
\caption{Examples of images masked with different strategies. Different masking strategies force the model to learn different representations in mask image modeling. 
}
\label{fig:ex_mask}
\end{figure}

\subsection{Multi-Masking Strategy}
\label{sec:framework} 

The pipeline of the proposed MMS is illustrated in Fig.~\ref{fig:arch}.
Following ViT \cite{vit} and MAE \cite{mae}, MMS first partitions the input text image $I$ into non-overlapping patches. Then, some patches are removed (masked) by random patch masking, block masking, and span masking in separate branches, while the remaining (visible) patches are fed into the model for reconstructing masked patches of each branch. The model in each branch shares the same weights.



\begin{figure}[!t]
\begin{algorithm}[H]
    \caption{Span Masking}
    \label{alg:span masking}
    \begin{algorithmic}
    \REQUIRE $N=(h \times w)$ image patches, masking ratio $R$, maximum masked span length $S$
    \ENSURE Masked positions $M$
    \STATE $M \leftarrow \emptyset$
    \REPEAT
    \STATE $s \leftarrow Rand(1, S)$
    \STATE $l \leftarrow Rand(0, max(0, N-s))$
    \STATE $r \leftarrow l + s$
    \IF{$R \leq 0.4$}
    \STATE $k \leftarrow s$
    \COMMENT{$k$:spacing}
    \ELSIF{$R \leq 0.7$}
    \STATE $k \leftarrow 1$
    \ELSE
    \STATE $k \leftarrow 0$
    \ENDIF
    \IF{$M \bigcap \{(i,j): i \in \{l-k,...,l-1\}, j \in \{0,...,h\}\} = \emptyset$ \AND
    $M \bigcap \{(i,j): i \in \{r+1,...,r+k\}, j \in \{0,...,h\}\} = \emptyset$}
    \STATE $M \leftarrow M \bigcup \{(i,j): i \in \{l,...,r\}, j \in \{0,...,h\}\}$
    \ENDIF
    \UNTIL{$|M| > R \cdot N$}
    \RETURN M
    \end{algorithmic}
\end{algorithm}
\end{figure}

\subsubsection{Masking}
The random patch masking branch follows MAE \cite{mae}, where image patches are randomly masked based on uniform random sampling with a certain masking ratio. 
The blockwise masking strategy, proposed in BEiT\cite{beit}, generates masks consisting of several rectangular blocks with random block sizes and aspect ratios, where these blocks are allowed to overlap. 
The span masking was originally introduced in PIXEL \cite{pixel} for sentence-level rendered text images, which are divided into single-row patch sequences. In this study, we expand it for word-level scene-text images with multi-row patches. By setting the mask ratio and maximum span length, span masking generates masks that completely cover all patches of some consecutive columns, resulting in the removal of the entire or large parts of individual characters. 
Algorithm~\ref{alg:span masking} details the generation process of the span masks and some examples of masked images are shown in Fig.~\ref{fig:ex_mask}.
Intuitively, random patch masking lets the model fill in part of characters from known pieces of characters, whereas blockwise and span masking challenge the model to predict the characters from the neighboring known characters or pieces. Consequently, blockwise and span masking promote a higher level of abstraction compared to random patch masking.
The hyperparameters, such as mask ratio and maximum span width, are investigated in the section \ref{MMS Analysis and Ablation Study}.
Here, the input image $I$ is divided and masked by random masking, block masking, and span masking, and the sets of visible patches are represented as $x_{r}$, $x_{b}$, and $x_{s}$, respectively.


\subsubsection{Encoder}
The encoder $F(\cdot)$ utilizes ViT, processing only the visible patches $x_{r}$, $x_{b}$, and $x_{s}$ from each branch to generate encoder features $f_{r}$, $f_{b}$, and $f_{s}$, respectively.
Given that the computational complexity of Transformer \cite{Vaswani2017transformer} increases quadratically with the number of tokens (patches), particularly in MMS, where the number of patches significantly multiplies due to multiple masking strategies, focusing solely on visible patches can accelerate the learning process and decrease the computational memory usage.

\subsubsection{Decoder}
The decoder reconstructs the images from the encoder output and mask tokens.
The augmented features ${f}^{\prime}_{r}$, ${f}^{\prime}_{b}$, and ${f}^{\prime}_{s}$, which are obtained by inserting mask tokens into $f_{r}$, $f_{b}$, and $f_{s}$, are fed into the decoder $R(\cdot)$ to produce reconstructed images $I_{r}$, $I_{b}$, and $I_{s}$, respectively.



\subsubsection{Loss Function}
Our reconstruction target is to predict the pixel values for each masked patch in every masking branch.  
Hence we minimize the mean-squared error (MSE) **only on the masked patches** of each branch.
Let $\mathcal{M}_r$, $\mathcal{M}_b$, and $\mathcal{M}_s$ denote the index sets of masked patches for the random, block, and span masking strategies, respectively. The branch-wise reconstruction losses are:

\begin{equation}
\label{eq:branch_loss}
\begin{aligned}
L_{r} &= \frac{1}{|\mathcal{M}_r|}\sum_{i \in \mathcal{M}_r} 
        \bigl\lVert I_{r,i} - I_{i}\bigr\rVert_{2}^{2},\\[2pt]
L_{b} &= \frac{1}{|\mathcal{M}_b|}\sum_{i \in \mathcal{M}_b} 
        \bigl\lVert I_{b,i} - I_{i}\bigr\rVert_{2}^{2},\\[2pt]
L_{s} &= \frac{1}{|\mathcal{M}_s|}\sum_{i \in \mathcal{M}_s} 
        \bigl\lVert I_{s,i} - I_{i}\bigr\rVert_{2}^{2},
\end{aligned}
\end{equation}
where $I_{r,i}$, $I_{b,i}$, and $I_{s,i}$ are the reconstructed pixel vectors of patch $i$ in each branch,
and $I_{i}$ is the corresponding normalized ground-truth patch.

Finally, the total multi-masking self-supervised loss is

\begin{equation}
L_{\text{MMS}} \;=\; L_{r} + L_{b} + L_{s}.
\end{equation}


\subsection{Text Recognition}
\label{sec:fine-tuning}
Our text recognition model, which follows that of DiG \cite{dig}, consists of a ViT encoder and a Transformer decoder.
The Transformer decoder comprises a 6-layer Transformer block and a fully connected layer for character prediction.
We adopt cross-entropy loss during the training.

\subsection{Text Super-resolution}
\label{sec:text sr}
Text super-resolution is the task of predicting high-resolution text images from low-resolution text images.
For this task, we employ a model composed of a ViT and a text super-resolution decoder.
The text super-resolution decoder consists of a 3-layer transformer block and a linear prediction head that predicts RGB pixel values.
The head number of the decoder is 2, and the embedding dimension is 384. For this task, L2 loss is employed.

\subsection{Text Segmentation}
\label{sec:text segmentation}
Text segmentation is a task that performs pixel-level binary classification of text foreground and background.
For text segmentation, we employ a model composed of a ViT encoder and a text segmentation decoder.
The decoder for text segmentation consists of a 3-layer transformer block and a linear prediction head for pixel class prediction.
The number of heads is set to 2 and the embedding dimension is 384.
We use cross-entropy loss for text segmentation.

\section{Experiment}
\label{Experiment}
\subsection{Dataset}
\textbf{Unlabeled Real Data (URD)} is an unlabeled real-world dataset comprising 15.77M images.
The text images are obtained from the OCR results of the Conceptual Caption Dataset
\footnote{\url{https://github.com/google-research-datasets/conceptual-captions}} by Microsoft Azure OCR.

\textbf{Synthetic Text Data (STD)} is a dataset consisting of 17M synthetic text images.
It is a combination of Synth90k\cite{mj} (9M) and SynthText\cite{gupta_synthetic_2016} (8M).

\textbf{Annotated Real Data (ARD)} is a labeled dataset containing 2.78M real-world images.
Images and labels are collected from TextOCR\cite{textocr} (0.71M) and Open Images Dataset v5\footnote{\url{https://storage.openvinotoolkit.org/repositories/openvino_training_extensions/datasets/open_images_v5_text}} (2.07M).

\textbf{Scene Text Recognition Benchmarks}
We assess the performance of the text recognition model with 11 scene text recognition benchmarks, classified into three categories: regular, irregular, and occluded datasets based on text complexity and layout.
The regular dataset contains IIIT5K-Words (IIIT) \cite{iiit}, Street View Text (SVT) \cite{SVT} and ICDAR2013 (IC13) \cite{KaratzasICDAR2013}, where text images with evenly spaced characters arranged horizontally.
Conversely, the irregular dataset includes ICDAR2015 (IC15) \cite{KaratzasICDAR2015}, SVT Perspective (SP) \cite{sp}, CUTE80 (CT) \cite{ct}, COCOText-Validation (COCO) \cite{veit_cocotext_2016}, CTW dataset \cite{ctw}, and Total-Text dataset (TT) \cite{tt}, which present challenging scenarios such as curved, rotated, or distorted text.
On the other hand, the occluded dataset, Weakly occluded scene text (WOST) \cite{ost} and Heavily occluded scene text (HOST) \cite{ost} were utilized to reflect the ability to recognize cases with missing visual cues. Images in this dataset are manually occluded in a weak or heavy degree.

\textbf{TextSeg} \cite{textseg} consist of 4024 annotated images for text segmentation.
Since this study focuses on cropped text images, we preprocessed the images and masks using word-level bounding boxes.
After preprocessing, there are 10421 training images and 3937 test images.

\textbf{TextZoom}\cite{Wangtextzoom2020} comprises pairs of high-resolution and low-resolution images for text super-resolution.
The training set includes 17367 image pairs, while the evaluation set is divided into three levels of difficulty: easy, medium, and hard, with 1619, 1411, and 1343 image pairs, respectively.

\subsection{Implementation Details}

\subsubsection{Self-supervised pre-training}
URD and STD were used as pre-training datasets, with input images of dimensions $32 \times 128$, and a patch size of $4 \times 4$. 
In our experiments, ViT-tiny, ViT-Small, and ViT-Base models were utilized as encoders. 
To streamline ablation studies and model analysis, we standardized ViT-Tiny as the default encoder to reduce the evaluation overhead.
For computational efficiency, a decoder with a depth of 2 and a dimension of 256 was employed.
The batch size was set at 512.
AdamW served as the optimizer with a $\beta$ of (0.9, 0.95) and a cosine learning schedule. The learning rate started at 1e-3 with 0.05 weight decay.
The warm-up was 5000 steps and the training epoch was set to 3 for ViT-Tiny and 10 for ViT-Small and ViT-Base models.


\subsubsection{Text Recognition Fine-Tuning}
During fine-tuning, the image size and patch size remained consistent with the pre-training phase. The training datasets employed were either STD or ARD. 
A batch size of 224 was utilized, along with the same optimizer and scheduler utilized during pre-training.
We used the same data augmentation method used in ABINet\cite{abinet}.
The default training epoch was 10 for the ablation study, extended to 35 solely for comparing the fine-tuning results with existing methods on ARD.
The base learning rate was set to 1e-4 with 0.05 weight decay. $\beta_1=0.9$, $\beta_2=0.999$, and the model warmed up with one epoch.
Evaluation was conducted on text recognition benchmarks with Top1 Accuracy serving as the evaluation metric.

\subsubsection{Text Segmentation}
Text segmentation shares the same image and patch sizes with text recognition.
The batch size is 256, with 800 training epochs and 50 warm-up epochs. 
Both the learning rate and optimizer settings align with those used for text recognition.
Evaluation is based on the Intersection over Union (IoU) metric.

\subsubsection{Text Super-resolution}
The setting for text super-resolution is the same as that for text recognition.
The batch size is 256, with 800 training epochs and 100 warm-up epochs.
Evaluation metrics include Peak Signal-to-Noise Ratio (PSNR) and Structure Similarity Index Measure (SSIM) \cite{wangSSIM2004}.

\begin{table}[t]
\centering
\caption{Ablation study on the masking strategies of MMS with different masking ratios.
Avg. is the per-image accuracy accuracy on all scene text recognition benchmarks.}
\label{tab:mms masking strategy}
\begin{tabular}{@{}ccc|c@{}}
\toprule
Random & Block & Span & Avg.  \\ \midrule
50\%   & -     & -    & 77.7             \\
75\%   & -     & -    & 77.8             \\
-      & 50\%  & -    & 79.3             \\
-      & 75\%  & -    & 77.7             \\
-      & -     & 50\% & 79.4             \\
-      & -     & 75\% & 77.2             \\ \midrule
75\%   & 50\%  & -    & 80.7             \\
75\%   & 75\%  & -    & 79.7             \\
75\%   & -     & 50\% & 79.4             \\
75\%   & -     & 75\% & 79.2             \\ \midrule
75\%   & 50\%  & 50\% & \textbf{81.2}    \\
75\%   & 50\%  & 75\% & 77.6             \\
75\%   & 75\%  & 50\% & 80.4             \\
75\%   & 75\%  & 75\% & 80.4             \\ \bottomrule
\end{tabular}
\end{table}

\begin{table}[t]
\centering
\caption{Survey on span width. Experiments were conducted only using span masking as a masking strategy in MAE. The evaluation metric is the top1 per-image accuracy on all scene text recognition benchmarks.}
\label{tab:span length}
\begin{tabular}{@{}ccccc@{}}
\toprule
\multirow{2}{*}{Masking Strategy} & \multicolumn{4}{c}{Span Length}    \\ \cmidrule(l){2-5} 
                                  & S=6  & S=8           & S=10 & S=12 \\ \midrule
Span Masking (50\%)               & 78.7 & \textbf{79.4} & 78.8 & 78.4 \\ \bottomrule
\end{tabular}
\end{table}

\subsection{Ablation Study and MMS Analysis}
\label{MMS Analysis and Ablation Study}
In this section, we first investigated the hyperparameters of MMS, including the combination of different mask strategies, the mask ratios of each masking branch, and the maximum span width in span masking. Subsequently, we discussed the effectiveness of random patch masking, block masking, and span masking strategies by comparing MMS with MAE variants with a single masking strategy. Our analysis delved into the representation extraction ability of different branches through fine-tuning evaluation, reconstruction quality evaluation, and attention visualization.


\subsubsection{Masking Strategy in MMS}
In this experiment, we first pre-trained each model 3 epochs with URD and STD, and then fine-tuned them 10 epochs with the ARD. Table~\ref{tab:mms masking strategy} shows the results derived from combining various masking strategies implemented in MMS. Considering the computation time and memory efficiency, we used 50\% and 75\% as the masking ratios. 
From the top section of Table~\ref{tab:mms masking strategy}, we first fixed the mask ratio of random masking in MMS at 75\%, which aligned with the MAE findings. 
Then, the results in the middle section of Table~\ref{tab:mms masking strategy} show that the results of models with two masking strategies were superior to those of a single masking strategy.
In particular, the average accuracy of the model trained with the combination of random (75\%) and block (50\%) masking improved that of block (50\%) masking by 1.4\%.
These findings underscore that models leveraging multiple masking strategies can effectively learn features crucial for text recognition. Further analysis of the bottom section of the table revealed that an ensemble of three masking strategies yielded superior accuracy compared to dual-strategy models.
Specifically, the model trained with the combination of random (75\%), block (50\%), and span (50\%) achieved the highest average accuracy, surpassing the dual-strategy model's accuracy by 0.5\% and the single-strategy model's accuracy by 1.8\%. 

These results highlight that the integration of three distinct masking strategies facilitated feature extraction across a broader spectrum of expression levels, thereby enhancing text recognition accuracy. Conversely, certain combinations harmed text recognition outcomes, emphasizing the significance of judiciously selecting the appropriate combination and mask ratios.
For the later experiment, MMS employed random (75\%), block (50\%), and span (50\%) as the default setting.

\subsubsection{Maximum Span Width of Span Masking}
We conducted experiments by varying the maximum width of the span from 6, 8, 10, to 12, while maintaining the mask ratio at 50\%.
The results are presented in Table~\ref{tab:span length}. 
The highest accuracy rate was obtained when the span width was set to 8.
Despite the marginal impact of span width, selecting an appropriate width still affects the performance.
We consider the results may be related to the average width of the characters in text images in the pre-training dataset and it should be reassessed when using it for other text images, such as handwritten images.

\begin{table}[t]
\centering
\caption{The comparison results between MMS and MAE with various masking strategies and ratios. Scratch is a model without pre-training. Avg. is the per-image accuracy on all scene text recognition benchmarks.}
\label{tab:mae fine-tuning}
\begin{tabular}{@{}lc@{}}
\toprule
\multicolumn{1}{c}{Method} & Avg. \\ \midrule
Scratch                    & 74.3 \\ \midrule
MAE (random 25\%)          & 76.3 \\
MAE (random 50\%)          & 77.7 \\
MAE (random 75\%)          & 77.8 \\ \midrule
MAE (block 25\%)           & 79.8 \\
MAE (block 50\%)           & 79.3 \\
MAE (block 75\%)           & 77.7 \\ \midrule
MAE (span 25\%)            & 78.2 \\
MAE (span 50\%)            & 79.4 \\
MAE (span 75\%)            & 77.2 \\ \midrule
\rowcolor[gray]{0.9}MMS    & \textbf{81.2} \\ \bottomrule
\end{tabular}
\end{table}

\subsection{Comparison with MAE}

\subsubsection{fine-tuning evaluation}
In this section, we conducted a comparative analysis among different models: the model without pre-training (referred to as Scratch), MAE models solely trained with one of the masking strategies (MAE random, MAE block, and MAE block), and MMS models, aiming to verify the effectiveness of employing multiple masking techniques.
Table \ref{tab:mae fine-tuning} presents the text recognition results of each model fine-tuned with ARD.
MMS achieved the highest average accuracy, showing a 6.9\% improvement over the scratch model.
Moreover, compared to MAE models utilizing random, block, and span masking, MMS exhibited average accuracy improvements of 3.4\%, 1.4\%, and 1.8\%, respectively.
These results demonstrate the superiority of MMS, which integrates multiple masking strategies, over MAE and its derivatives utilizing single masking strategies

\begin{table}[t]
\centering
\label{tab:reconstruction}
\caption{Qualitative evaluation of the model trained with different masking strategies on evaluation dataset with different masking methods. The best PSNR values are in bold and the second best values are underlined.}

\begin{tabular}{@{}ccccc@{}}
\toprule
\multirow{2}{*}{Methods} & \multicolumn{3}{c}{Evaluation Sets}              & \multirow{2}{*}{Avg.} \\ \cmidrule(lr){2-4}
                         & random 75\%    & block 50\%     & span 50\%      &                      \\ \midrule
MAE (random 75\%)        & \textbf{29.02} & 25.07 & 26.71 & 26.94                \\
MAE (block 50\%)         & 28.27 & \textbf{26.29} &  \underline{27.67} & \textbf{27.41}                \\
MAE (span 50\%)          & 23.03 & 23.05 & \textbf{28.15} & 24.74                \\
MMS                      & \underline{28.29} & \underline{25.87} & \underline{27.67} & \underline{27.28}       \\ \bottomrule
\end{tabular}
\end{table}

\begin{figure}[t]
\centering
\includegraphics[width=1.0\columnwidth]{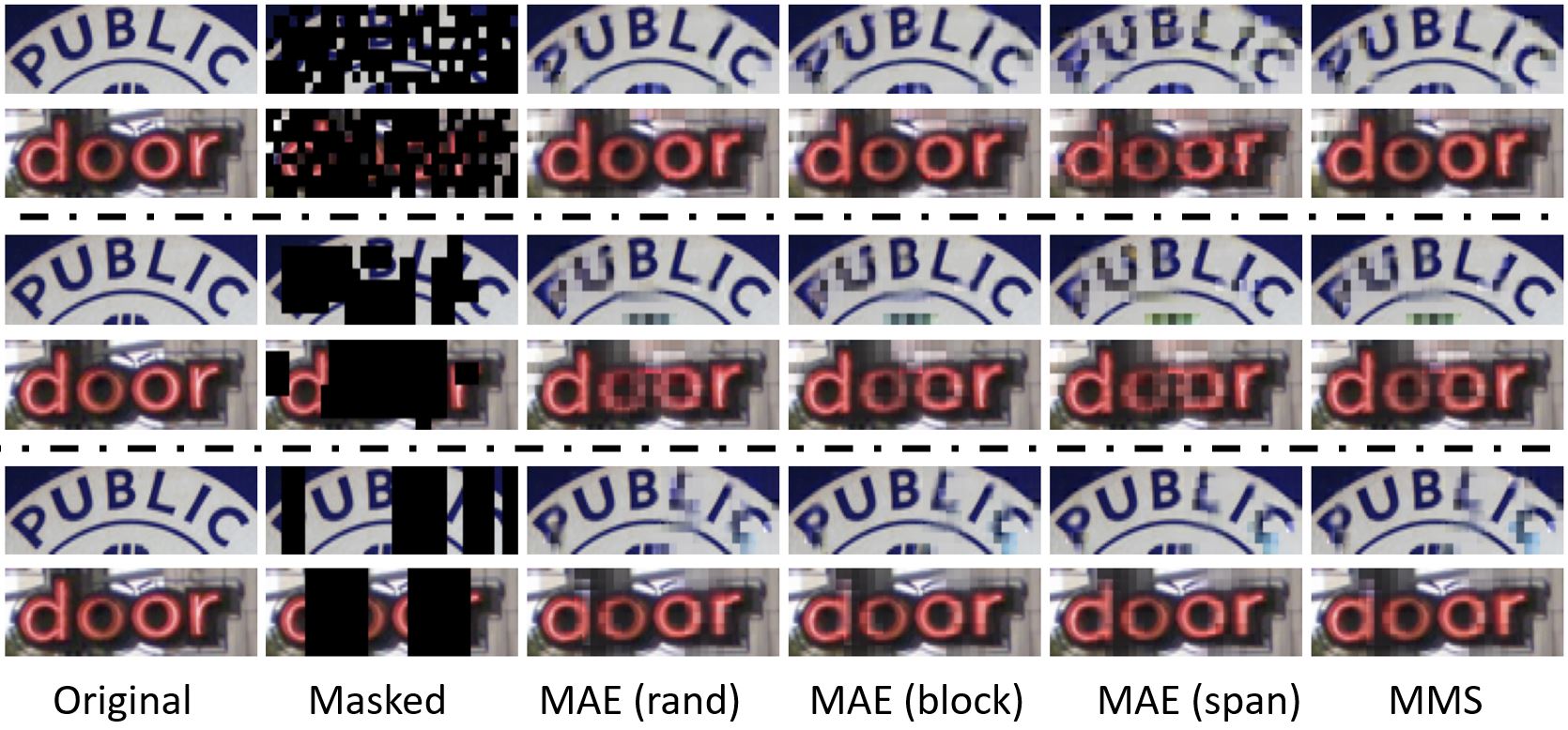}
\caption{Reconstructions of scene text benchmarks images. From left to right: original image, masked image (top: random masking (75\%); middle: block masking (50\%); bottom: span masking (50\%)), images reconstructed by MAE (random75\%), images reconstructed by MAE (block 50\%), images reconstructed by MAE (span 50\%), images reconstructed by MMS.}
\label{fig:reconstruction}
\end{figure}

\subsubsection{Reconstruction of Masked Images}
We conducted a qualitative evaluation experiment on the models pre-trained with MAE and MMS to assess the quality of images reconstructed from masked images.
We first created three evaluation datasets utilizing random patch masking (75\%), block masking (50\%), and span masking (50\%) on the IIIT dataset. Then we evaluated the quality of reconstructed images of different models across these datasets using PSNR metrics. The PSNR was calculated between the original image and an image in which only the masked portion was replaced by the prediction result.
The results are presented in Table~\ref{tab:reconstruction}.
Within each evaluation set, the MAE model trained with the corresponding masking strategy achieved the highest PSNR. Conversely, other MAE models trained with different masking strategies had a significant performance decline. Specifically, in the random 75\% set and span 50\% set, the performance gap between MAE (random 75\%) and MAE (span 50\%) were 5.99 dB and 1.44 dB, respectively. Both MAE (random 75\%) and MAE (span 50\%) performed poorly in the block 55\% set, while MAE (block 50\%) achieved a relatively balanced performance across both evaluation sets. We speculate that block masking yields both small scattered and large consecutive masking regions, leading to an intermediate state between random patch masking and span masking. 
These results suggest disparities in data generated with different masking strategies and that different masking strategies can compel MAE to learn distinct representations. 
On the other hand, our MMS consistently obtained the second-highest PSNR in each evaluation set, indicating its proficiency in reconstructing diverse masked data well and learning comprehensive representations from varied masking strategies simultaneously. 
Some examples of the reconstructed image are depicted in Fig.~\ref{fig:reconstruction}.
MAEs trained with a single masking strategy tend to generate blurry and incorrect content when encountering images masked with differing strategies. In contrast, MMS consistently delivers clear and correct reconstruction results in each evaluation set.

\begin{figure}[t]
\centering
\includegraphics[width=\linewidth]{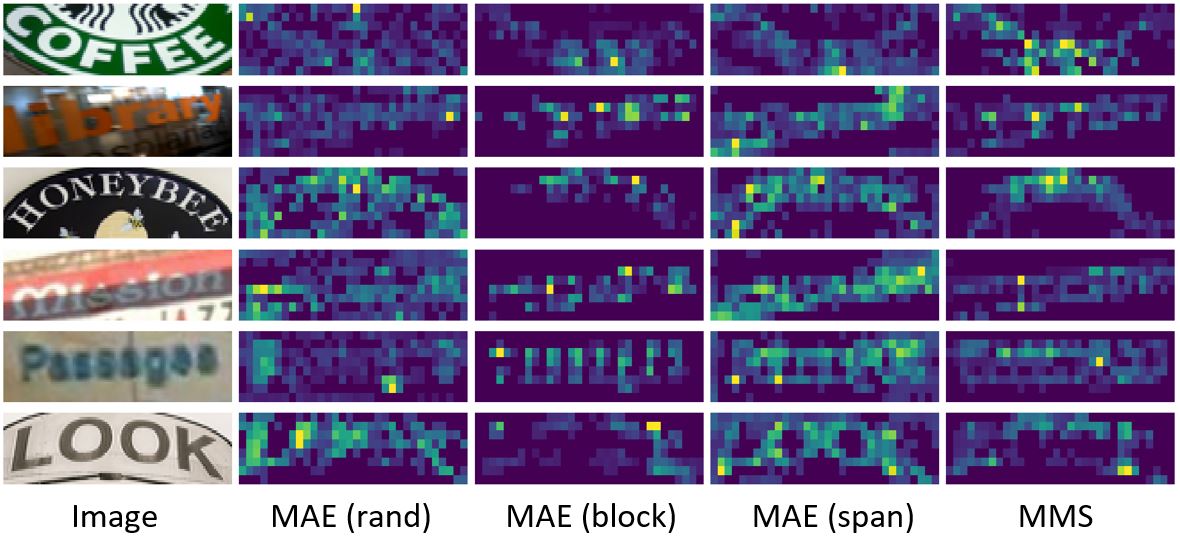}
\caption{Visualization results of the attention map of the [CLS] token.} 
\label{fig:attention_cls}
\end{figure}
\begin{figure*}[t]
\centering
\includegraphics[width=\linewidth]{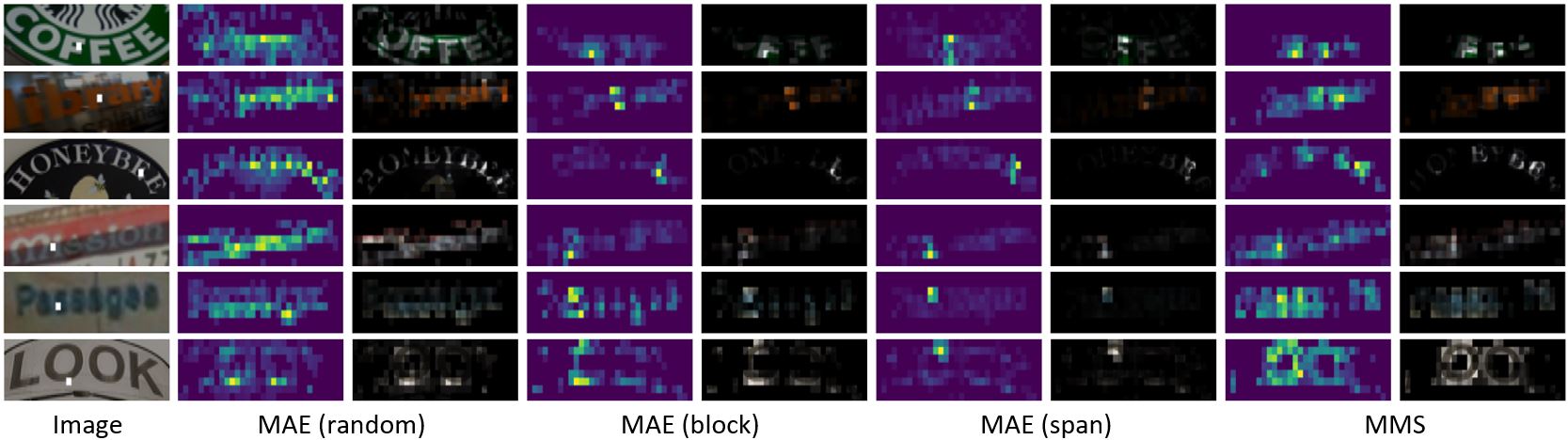}
\caption{Visualization results of the attention map corresponding to the specified patch. The specified patches are shown in white in the first image. The second images of each image pair are made by masking the original image with black using attention value. Areas with higher attention values are more transparent.}
\label{fig:attention_patch}
\end{figure*}
\begin{figure*}[t]
\centering
\includegraphics[width=\linewidth]{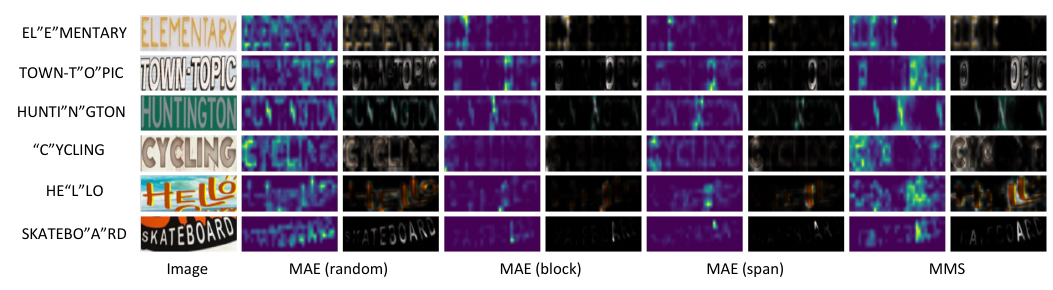}
\caption{Visualization results of the attention map corresponding to the character instances.
The specified characters are enclosed in double quotation marks.}
\label{fig:attention_char}
\end{figure*}
\begin{figure}[t]
\centering
\includegraphics[width=\linewidth]{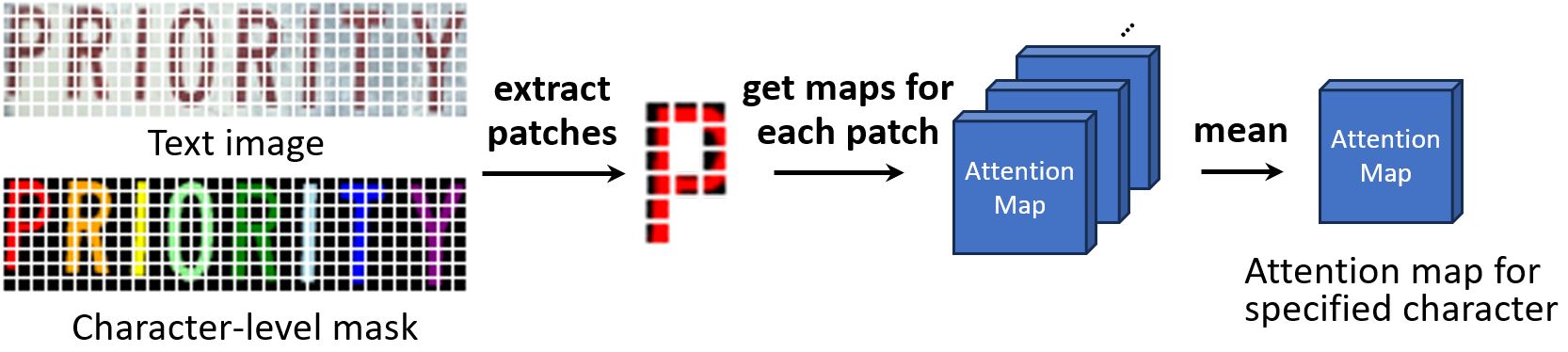}
\caption{How to get an attention map for a specified character instance.
First, image patches are extracted in which more than 70\% of the pixels overlap with the segmentation map for the specified character. Then, attention maps are obtained for each image patch. Finally, the obtained attention maps are averaged and visualized.}
\label{fig:attention_char_method}
\end{figure}
\begin{figure}[t]
\centering
\includegraphics[]{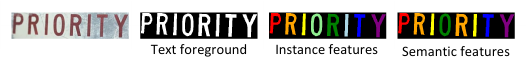}
\caption{Three types of text feature representations, where the semantic feature is considered more useful for the text recognition task.}
\label{fig:features}
\end{figure}

\begin{table*}[t]
\centering

\tabcolsep 3pt
\caption{Comparison results with existing self-supervised text recognition methods.\\
Avg1 is the weighted average accuracy of IIIT, SVT, IC13, IC15, SP, and CT by size.\\
Avg2 is the weighted average accuracy of all benchmarks by size.}
\label{tab:self-supervised comparisons}
\begin{tabular}{lcccccccccccccc}
\toprule
\multicolumn{1}{c}{\multirow{2}{*}{Method}} & \multirow{2}{*}{Data} & \multicolumn{3}{c}{Regular} & \multicolumn{6}{c}{Irregular} & \multicolumn{2}{c}{Occluded} & \multirow{2}{*}{Avg1} & \multirow{2}{*}{Avg2}\\
\cmidrule(lr){3-5} \cmidrule(lr){6-11} \cmidrule(lr){12-13}
&& IIIT & SVT & IC13 & IC15 & SP & CT & COCO & CTW & TT & HOST & WOST \\
\midrule
 SeqCLR\cite{seqclr}& STD & 82.9 & - & 87.9 & - & - & - & - & - & - & - & - & - & -\\
 PerSec-ViT + UTI-100M\cite{persec} & STD & 88.1 & 86.8 & 94.2 & 73.6 & 77.7 & 72.7 & - & - & - & - & - & 83.77 & -\\
\midrule
 DiG (ViT-Tiny)\cite{dig} & STD & 95.8 & 92.9 & 96.4 & 84.8 & 87.4 & 86.1 & 66.8 & 75.3 & 78.1 & 60.9 & 73.0 & 91.83 & \underline{75.46} \\
 CCD (ViT-Tiny)\cite{ccd} & STD & 96.5 & 93.4 & 96.3 & 85.2 & 89.8 & 89.2 & - & - & - & - & - & \textbf{92.57} & -\\
 \rowcolor[gray]{0.9}
 MMS (ViT-Tiny) (Ours) & STD & 95.7 & 93.7 & 94.7 & 85.4 & 87.4 & 89.6 & 66.1 & 76.0 & 79.2 & 64.8 & 75.8 & \underline{91.91} & \textbf{75.97} \\
\midrule
 DiG (ViT-Small)\cite{dig} & STD & 96.7 & 93.4 & 97.1 & 87.1 & 90.1 & 88.5 & 68.8 & 78.8 & 81.1 & 72.1 & 81.1 & \underline{93.23} & \textbf{78.89} \\
 MaskOCR (ViT-Small)\cite{maskocr} & STD & 95.8 & 94.0 & 97.7 & 87.5 & 90.2 & 89.2 & - & - & - & - & - & 93.0 & -\\
 CCD (ViT-Small)\cite{ccd} & STD & 96.8 & 94.4 & 96.6 & 87.3 & 91.3 & 92.4 & - & - & - & - & - & \textbf{93.59} & -\\
 \rowcolor[gray]{0.9}
 MMS (ViT-Small) (Ours) & STD & 96.7 & 94.0 & 95.2 & 86.8 & 88.2 & 91.0 & 68.6 & 77.2 & 81.3 & 69.0 & 79.8 & 92.87 & \underline{78.23}\\
\midrule
 DiG (ViT-Base)\cite{dig} & STD & 96.7 & 94.6 & 96.9 & 87.1 & 91.0 & 91.3 & 69.8 & 79.3 & 81.9 & 74.9 & 82.3 & \underline{93.49} & \textbf{79.78}\\
 MaskOCR (ViT-Base)\cite{maskocr} & STD & 95.8 & 94.9 & 98.1 & 87.5 & 89.8 & 90.3 & - & - & - & - & - & 93.1 & -\\
 CCD (ViT-Base)\cite{ccd} & STD & 97.2 & 94.4 & 97.0 & 87.6 & 91.8 & 93.3 & - & - & - & - & - & \textbf{93.96} & -\\
 \rowcolor[gray]{0.9}
 MMS (ViT-Base) (Ours) & STD & 96.7 & 94.2 & 95.4 & 87.0 & 89.8 & 91.3 & 68.9 & 78.9 & 82.8 & 73.4 & 81.5 & 93.12 & \underline{79.20} \\
\midrule
\midrule
 DiG (ViT-Tiny)\cite{dig} & ARD & 96.4 & 94.4 & 96.2 & 87.4 & 90.2 & 94.1 & 71.8 & 83.1 & 86.6 & 45.3 & 68.2 & 93.37 & \underline{77.10} \\
 CCD (ViT-Tiny)\cite{ccd} & ARD & 97.1 & 96.0 & 97.5 & 87.5 & 91.6 & 95.8 & - & - & - & - & - & \underline{94.18} & -\\
 \rowcolor[gray]{0.9}
 MMS (ViT-Tiny) (Ours) & ARD & 98.0 & 97.6 & 97.7 & 89.4 & 93.9 & 96.2 & 77.2 & 88.1 & 91.3 & 68.5 & 81.1 & \textbf{95.36} & \textbf{83.79} \\
\midrule
 DiG (ViT-Small)\cite{dig} & ARD & 97.7 & 96.1 & 97.3 & 88.6 & 91.6 & 96.2 & 75.0 & 86.3 & 88.9 & 56.0 & 75.7 & 94.69 & 80.79\\
 MaskOCR (ViT-Small)\cite{maskocr} & ARD & 98.0 & 96.9 & 97.8 & 90.2 & 94.9 & 96.2 & - & - & - & - & - &  \underline{95.6} & -\\
 CCD (ViT-Small)\cite{ccd} & ARD & 98.0 & 96.4 & 98.3 & 90.3 & 92.7 & 98.3 & 76.7 & 86.5 & 91.3 & 77.3 & 86.0 &95.57 & \underline{84.85} \\
 \rowcolor[gray]{0.9}
 MMS (ViT-Small) (Ours) & ARD & 98.2 & 98.0 & 98.2 & 90.4 & 94.1 & 96.9 & 78.7 & 88.9 & 92.5 & 74.3 & 83.7 & \textbf{95.85} & \textbf{85.45}\\
\midrule
 DiG (ViT-Base)\cite{dig} & ARD & 97.6 & 96.5 & 97.6 & 88.9 & 92.9 & 96.5 & 75.8 & 87.0 & 90.1 & 62.8 & 79.7 & 94.92 & \underline{82.31}\\
 MaskOCR (ViT-Base)\cite{maskocr} & ARD & 98.0 & 96.9 & 98.2 & 90.1 & 94.6 & 95.8 & - & - & - & - & - & 95.6 & -\\
 CCD (ViT-Base)\cite{ccd} & ARD & 98.0 & 97.8 & 98.3 & 91.6 & 96.1 & 98.3 & - & - & - & - & - & \textbf{96.30} & -\\
 \rowcolor[gray]{0.9}
 MMS (ViT-Base) (Ours) & ARD & 98.1 & 97.0 & 98.6 & 91.0 & 96.3 & 97.6 & 79.9 & 88.4 & 92.9 & 78.6 & 86.2  & \underline{96.17} & \textbf{86.62} \\
\bottomrule
\end{tabular}
\end{table*}

\subsubsection{Attention analysis}
In this section, we visualized and analyzed the attention map by inputting the text image into the pre-trained encoder of various pretrained methods to investigate the activated latent representations. We compared MAEs with random patch masking (75\%), block masking (50\%), span masking (50\%), and MMS by visualizing three distinct types of attention maps: the [CLS] token, the specified patch, and the specified text instance. 
For the first two types of attention maps, we followed the attention heatmap visualization in DINO \cite{dino}, where the attention weights in the final layer of the encoder were averaged over all heads and thresholded for clear visualization. The attention visualization of the specified text instance will be discussed later.

The [CLS] token is regarded as an aggregated representation of the entire image, specifically designed for classification purposes in ViT models. Although the [CLS] token is not used for fine-tuning purposes in this study, it can be considered as compact information regarding the image for reference. 
The attention maps on the [CLS] tokens are depicted in Fig.~\ref{fig:attention_cls}.
The attention patterns of MAE pre-trained with random masking are activated holistically rather than solely concentrating on the text regions, resulting in a dispersion of attention. Block masking leads to the loss of attention towards some distant characters. In contrast, models trained with span masking and MMS direct their attention towards the text, with MMS exhibiting a more succinct capture of information compared to span masking.
Therefore, MMS is deemed proficient in grasping concise yet vital information for text image recognition. 


Subsequently, we plot the attention map corresponding to the specified patch in Fig.~\ref{fig:attention_patch}.
Given a patch containing text pixels, the model pre-trained with random patch masking focuses on the entire foreground region of the text, highlighting adjacent characters in the images as well. This observation implies that random patch masking enables the model to learn relatively low-level stroke information, unable to separate features from different characters.
On the other hand, MAEs pre-trained with block masking and span masking focus on the precise character region that includes the specified patch. This suggests that block masking and span masking can distinguish character-level features from text images. Regarding MMS, the attention maps not only focus on the character region containing the specified patch but also emphasize the same character within the text images. This finding underscores that MMS allows the model to glean character-level and stroke-level features through different masking branches.


Finally, we create attention maps for the specified character instance. Fig.~\ref{fig:attention_char_method} illustrates the generation process of these maps. Initially, we utilize the text mask in the TextSeg dataset to pick up the patches whose areas are occupied by text pixels more than 70\%. Following this, we collected the attention maps associated with the selected patches and averaged them to produce the attention map for the specified character instance.
The generated attention maps are displayed in Fig.~\ref{fig:attention_char}.
Broadly, in the attention maps of random masking, features related to the text foreground regions are holistically activated, with the specified character and its similar character having a higher attention value. In the case of block and span masking, the attention maps primarily concentrate on the region encompassing the specified character. Meanwhile, With MMS, both the regions and the regions containing the specified character and those with the same specified character are highlighted. For example, in the word "ELEMENTARY`` (first row), although the third letter "E" is specified, not only the regions of the specified third "E" but also those of the first and fifth letters "E" have higher attention values.
This observation is similar to the attention maps of the specified patch, which indicate random masking excels in yielding stroke-level features to separate text foreground from background, whereas block and span masking could capture character-level features to discriminate character instances. In addition, MMS not only extracts character-level features but also discerns the relationship among different character instances. 
CCD \cite{ccd} discussed different textual features in the self-supervised learning for text recognition, including text foreground, instance features, and semantic features, as depicted in Fig.~\ref{fig:features}. Among these types of features, semantic features pose a greater learning challenge but offer enhanced utility for the text recognition task. Through our analysis of the attention maps, we found that random patch masking learns text foreground features, block and span masking capture instance features, and MMS identifies semantic features from text images.

\begin{table*}[t]
\centering

\tabcolsep 3pt
\caption{Comparison results of MMS with existing self-supervised text recognition methods when training with different data ratios.}
\label{tab:fine-tuning data ratio}
\begin{tabular}{clcccccccccccc}
\toprule
\multirow{2}{*}{Label Fraction}& \multicolumn{1}{c}{\multirow{2}{*}{Method}} & \multicolumn{3}{c}{Regular} & \multicolumn{6}{c}{Irregular} & \multicolumn{2}{c}{Occluded} & \multirow{2}{*}{Avg.}\\
\cmidrule(lr){3-5} \cmidrule(lr){6-11} \cmidrule(lr){12-13}
& & IIIT & SVT & IC13 & IC15 & SP & CT & COCO & CTW & TT & HOST & WOST \\
\midrule
\multirow{3}{*}{1\%(27.8K)}
 & DiG-ViT-Small\cite{dig} & 88.4 & 86.2 & 89.9 & 79.0 & 76.6 & 77.8 & 54.8 & 67.9 & 67.2 & 33.2 & 53.3 & 62.9\\
 & CCD-ViT-Small\cite{ccd} & 89.3 & 86.5 & 88.8 & 76.5 & 80.1 & 74.7 & 54.9 & 65.5 & 67.8 & 38.4 & 55.9 & 63.7\\
 \rowcolor[gray]{0.9}
 \cellcolor[gray]{1}
 & MMS-ViT-Small (Ours) & 94.6 & 93.6 & 94.7 & 83.5 & 86.0 & 91.7 & 65.3 & 77.6 & 78.6 & 41.8 & 66.8 & \textbf{72.5}\\
\midrule
\multirow{3}{*}{10\%(278K)}
 & DiG-ViT-Small\cite{dig} & 95.3 & 94.4 & 95.9 & 85.3 & 87.9 & 91.7 & 67.1 & 80.5 & 81.1 & 42.1 & 64.0 & 73.5\\
 & CCD-ViT-Small\cite{ccd} & 95.9 & 94.1 & 96.6 & 87.1 & 89.9 & 94.1 & 69.2 & 81.6 & 84.3 & 63.4 & 76.2 & 78.2\\
 \rowcolor[gray]{0.9}
 \cellcolor[gray]{1}
 & MMS-ViT-Small (Ours) & 97.1 & 96.2 & 96.7 & 88.5 & 91.0 & 95.5 & 73.8 & 86.9 & 88.6 & 60.8 & 76.4 & \textbf{80.8}\\
\midrule
\multirow{3}{*}{100\%(2.78M)}
 & DiG-ViT-Small\cite{dig} & 97.7 & 96.1 & 97.3 & 88.6 & 91.6 & 96.2 & 75.0 & 86.3 & 88.9 & 56.0 & 75.7 & 80.7\\
 & CCD-ViT-Small\cite{ccd} & 98.0 & 96.4 & 98.3 & 90.3 & 92.7 & 98.3 & 76.7 & 86.5 & 91.3 & 77.3 & 86.0 & 84.9\\
 \rowcolor[gray]{0.9}
 \cellcolor[gray]{1}
 & MMS-ViT-Small (Ours) & 98.2 & 98.0 & 98.2 & 90.4 & 94.1 & 96.9 & 78.7 & 88.9 & 92.5 & 74.3 & 83.7 & \textbf{85.5}\\
\bottomrule
\end{tabular}
\end{table*}

\begin{table*}[t]
\centering
\tabcolsep 3pt
\caption{Feature representation evaluation of MMS on all scene text recognition benchmarks.}
\label{tab:feature representation evaluation}
\begin{tabular}{lcccccccccccc}
\toprule
\multicolumn{1}{c}{\multirow{2}{*}{Method}} & \multicolumn{3}{c}{Regular} & \multicolumn{6}{c}{Irregular} & \multicolumn{2}{c}{Occluded} & \multirow{2}{*}{Avg.}\\
\cmidrule(lr){2-4} \cmidrule(lr){5-10} \cmidrule(lr){11-12}
& IIIT & SVT & IC13 & IC15 & SP & CT & COCO & CTW & TT & HOST & WOST \\
\midrule
 Gen-ViT-Small\cite{dig} & 86.6 & 82.1 & 88.7 & 72.9 & 74.4 & 72.2 & 48.5 & 64.1 & 63.3 & 33.8 & 56.5 & 59.3\\
 Dis-ViT-Small\cite{dig} & 92.6 & 90.4 & 93.4 & 81.2 & 81.7 & 84.0 & 60.0 & 72.8 & 73.1 & 33.3 & 56.1 & 67.0\\
 DiG-ViT-Small\cite{dig} & 94.2 & 93.0 & 95.3 & 84.3 & 86.1 & 87.5 & 63.4 & 77.9 & 75.8 & 41.7 & 64.0 & 71.1\\
 CCD-ViT-Small\cite{ccd} & 93.5 & 89.6 & 92.8 & 82.7 & 85.1 & 83.0 & 60.4 & 73.3 & 73.4 & 47.6 & 66.5 & 69.9\\
 \rowcolor[gray]{0.9}
 MMS-ViT-Small (Ours) & 94.2 & 92.6 & 94.3 & 84.0 & 87.1 & 89.2 & 62.0 & 78.0 & 76.6 & 58.1 & 73.9 & \textbf{73.2}\\
\bottomrule
\end{tabular}
\end{table*}

\subsection{Comparison With State-of-the-Art Methods}

\subsubsection{Text Recognition}
\paragraph{Self-supervised text recognition}
\label{sec:self-supervised text recognition}

We compared our MMS with existing self-supervised text recognition methods, and the results are presented in Table~\ref{tab:self-supervised comparisons}. Compared to SeqCLR and PerSec, even our smallest MMS-ViT-Tiny significantly outperformed them in recognition accuracy across all datasets. Despite PerSec being pre-trained on 100 million images, the MMS-ViT-Series achieved performance gains of 8.14\%, 9.1\%, and 9.35\% respectively.
Furthermore, we conducted a comparison of MMS with state-of-the-art methods DiG and CCD, all utilizing the same text recognition network, pre-trained with URD and STD, and fine-tuned with STD or ARD. 
The top section of Table~\ref{tab:self-supervised comparisons} displays the result of text recognition networks using various ViT backbones and fine-tuned with STD. While MMS-ViT-Tiny outperforms DiG-ViT-Tiny on Avg1 and Avg2, it was inferior to CCD--ViT-Tiny. Additionally, MMS-ViT-Small and MMS-ViT-Base also underperformed compared to their DiG and CCD counterparts.

However, when fine-tuning with ARD, MMS-Series exhibited significantly better recognition performances than DiG-Series and CCD-Series. MMS-Series outperforms DiG-Series By 1.99\%, 1.16\%, and 1.25\% on Avg1 and by 6.69\%, 4.66\%, and 4.32\% on Avg2. Moreover, MMS-ViT-Tiny and MMS-ViT-Small surpass CCD-ViT-Tiny and CCD--ViT-Small by 1.18\% and 0.28\% on Avg1, respectively. 

It is noteworthy that MMS-ViT-Series achieves higher performance gains on Avg2 than Avg1, indicating their superior performance on curved text and occluded text datasets such as COCO, CTW, TT, HOST, and WOST. The complex layouts of curved text pose challenges for contrastive learning methods, whereas Mask Image Modeling (MIM) excels without the need for character discrimination and is adept at handling occluded text.
Additionally, the performance gains from MMS-ViT-Tiny to MMS-ViT-Base gradually decrease, possibly due to the consistent use of the same small reconstruction decoder across all MMS-Series models, impacting the pre-training performance of larger models like MMS-ViT-Base.
In summary, these results underscore the superiority of our proposed MIM paradigm over existing contrastive learning methods, particularly on real-world datasets.

\begin{table*}[t]
\centering

\tabcolsep 3pt
\caption{Comparison results with existing text recognition methods.
Type V and L denote models that use only visual models and models that use language models in addition to visual models, respectively.
Avg-IC13 is a weighted average of IIIT, SVT, IC13, SVTP, and CT by size.
Avg-IC15 is the weighted average of IIIT, SVT, IC15, SVTP, and CT by size.}
\label{tab:supervised comparison}
\begin{tabular}{l|c|c|cccccc|cc|c}
\toprule
Method & Type & Data & IIIT & SVT & IC13 & IC15 & SP & CT & Avg-IC13 & Avg-IC15 & Params.\\
\midrule
 SATRN\cite{satrn} & \multirow{5}{*}{V} & STD & 92.8 & 91.3 & - & - & 86.5 & 87.8 & - & - & 55M \\
 MGP-STR\cite{mgp-str} && STD & 96.4 & 94.7 & - & 87.2 & 91.0 & 90.3 & - & 92.80 & 148M \\
 SGBANet\cite{sgbanet} && STD & 95.4 & 89.1 & 95.1 & - & 83.1 & 88.2 & 92.83 & - & - \\
 CornerTransformer\cite{corner-transformer} && STD & 95.9 & 94.6 & 96.4 & - & 91.5 & 92.0 & 95.13 & - & 86M \\
 SIGA\cite{siga} && STD & 96.6 & 95.1 & 96.8 & 86.6 & 90.5 & 93.1 & \underline{95.58} & \underline{92.84} & 113M \\
\midrule
 ABINet\cite{abinet} & \multirow{5}{*}{L} & STD+WiKi & 96.2 & 93.5 & - & 86.0 & 89.3 & 89.2 & - & 92.02 & 37M \\
 S-GTR\cite{s-gtr} && STD+WiKi & 95.8 & 94.1 & - & 84.6 & 87.9 & 92.3 & - & 91.50 & 42M \\
 ABINet+ConCLR\cite{conclr} && STD+WiKi & 96.5 & 94.3 & - & 85.4 & 89.3 & 91.3 & - & 92.17 & -\\
 LevOCR\cite{levocr} && STD & 96.6 & 92.9 & - & 86.4 & 88.1 & 91.7 & - & 92.26 & 109M \\
 PARSeq\cite{parseq} && STD & 97.0 & 93.6 & 96.2 & 86.5 & 88.9 & 92.2 & {95.28} & {92.65} & - \\
\midrule
  MMS-ViT-Tiny  & \multirow{3}{*}{V} & STD & 95.7 & 93.7 & 94.7 & 85.4 & 87.4 & 89.6 & 93.71 & 91.08 & 20M \\
 MMS-ViT-Small  && STD & 96.7 & 94.0 & 95.2 & 86.8 & 88.2 & 91.0 & 94.84 & {92.51} & 36M \\
 MMS-ViT-Base && STD & 96.7 & 93.3 & 96.2 & 86.4 & 89.5 & 91.3 & {95.11} & 92.47 & 52M \\
\midrule
 MMS-ViT-Tiny  & \multirow{3}{*}{V} & ARD & 98.0 & 97.6 & 97.7 & 89.4 & 93.9 & 96.2 & 97.31 & 95.00 & 20M \\
 MMS-ViT-Small  && ARD &\bf{98.2} & \bf{98.0} & 98.2 & 90.4 & 94.1 & 96.9  & 97.64 & 95.50 & 36M \\
 MMS-ViT-Base && ARD & 98.1 & 97.0 & \bf{98.6} & \bf{91.0} & \bf{96.3} & \bf{97.6} & \bf{97.84} & \bf{95.78} & 52M \\
\bottomrule
\end{tabular}
\end{table*}

\begin{table}[t]
\centering
\tabcolsep 3pt
\caption{The super-resolution evaluation results on the TextZOOM benchmark.}
\label{tab:sr result}
\begin{tabular}{lcccccccc}
\toprule
\multirow{2}{*}{Method} &  \multicolumn{4}{c}{SSIM(\%)↑} & \multicolumn{4}{c}{PSNR↑}\\
\cmidrule(lr){2-5} \cmidrule(lr){6-9}
& Easy & Med & Hard & Avg. & Easy & Med & Hard & Avg. \\
\midrule
Bicubic &  78.84 & 62.54 & 65.92 & 69.61 & 22.35 & 18.98 & 19.39 & 20.35\\
SRCNN\cite{Dong2016SRCNN} & 83.79 & 63.23 & 67.91 & 72.27 & 23.48 & 19.06 & 19.34 & 20.78\\
SRResNet\cite{Ledig2017SRResNet} &  86.81 & 64.06 & 69.11 & 74.03 & 24.36 & 18.88 & 19.29 & 21.03\\
HAN\cite{Niu2020HAN} &  86.91 & 65.37 & 73.87 & 75.96 & 23.30 & 19.02 & 20.16 & 20.95\\
TSRN\cite{Wangtextzoom2020} &  88.97 & 66.76 & 73.02 & 76.90 & 25.07 & 18.86 & 19.71 & 21.42\\
TBSRN\cite{Chen2021TBSRN} &  87.29 & 64.55 & 74.52 & 76.03 & 23.46 & 19.17 & 19.68 & 20.91\\
PCAN\cite{Zhao2021PCAN} &  88.30 & 67.81 & 74.75 & 77.52 & 24.57 & 19.14 & 20.26 & 21.49\\
\midrule
Scratch-Small & 81.43 & 62.88 & 68.45 & 71.56 & 22.90 & 19.65 & 20.45 & 21.10\\
DiG-Small\cite{dig} & 86.13 & 65.61 & 72.15 & 75.22 & 23.98 & 19.85 & 20.57 & 21.60\\
CCD-Small\cite{ccd} & 88.22 & 70.05 & 75.43 & \bf{78.43} & 24.40 & 20.12 & 20.18 & \underline{21.84}\\
\rowcolor[gray]{0.9}
MMS-Small & 88.82 & 68.45 & 74.91 & \underline{77.98} & 25.29 & 20.41 & 21.10 & \bf{22.43}\\
\bottomrule
\end{tabular}
\end{table}


\begin{table}[t]
\centering
\tabcolsep 3pt
\caption{The text segmentation results on the TextSeg benchmark.}
\label{tab:segmentation result}
\begin{tabular}{ccccc}
\toprule
\multirow{2}{*}{Method} & Scratch- & DiG- & CCD- & MMS- \\
& ViT-Small & ViT-Small\cite{dig} & ViT-Small\cite{ccd} & ViT-Small (Ours) \\
\midrule
IoU(\%)↑ & 78.1 & 83.1 & 84.8 & \textbf{85.0}\\
\bottomrule
\end{tabular}
\end{table}

\paragraph{Fine-tuning with Different Data Ratios}
We conducted a comparative analysis between MMS with DiG and CCD using various data ratios to demonstrate the effectiveness of pre-training. Specifically, we fine-tuned the MMS-ViT-small using 1\%, 10\%, and 100\% of ARD. 
The evaluation results on text recognition benchmarks are shown in Table~\ref{tab:fine-tuning data ratio}.
Our proposed MMS-ViT-Small outperforms the state-of-the-art method CCD-ViT-Small by 6.6\%, 2.6\%, and 0.6\% when fine-tuned with 1\%, 10\%, and 100\% of ARD, respectively.
Notably, MMS outperforms other methods by a large margin when fine-tuning with only 1\% of ARD. This suggests that MMS effectively learns a robust textual representation from unlabeled data and can be easily adapted with a small amount of labeled data for text recognition tasks.


\paragraph{Feature Representation Evaluation}

Following DiG and CCD, we assessed the quality of pre-trained features by freezing the encoder's parameters of the text recognition model during fine-tuning, using ARD as the dataset.
The evaluation results on text recognition benchmarks are detailed in Table~\ref{tab:feature representation evaluation}.
MMS demonstrates superior performance over DiG and CCD, achieving average accuracy improvements of 2.1\% and 3.3\%, respectively. 
In general, discrimination pretext tasks typically focus on segregating character instances in latent space, which is advantageous for classification tasks such as text recognition.
While DiG and CCD employ contrastive learning for instance (character) discrimination, our MMS relies solely on image reconstruction as pretext tasks and MMS surpasses DiG and CCD in accuracy. This indicates that MMS learns high-quality features able to discriminate characters from MIM.
Notably, MMS enhances the previous leading model CCD by 10.5\% and 7.4\% on the WOST and HOST datasets, respectively, due to the similar appearance of occluded images and masked images.



\paragraph{Scene Text Recognition}

We compared the proposed MMS with existing supervised text recognition methods in Table~\ref{tab:supervised comparison}. When the models were trained with STD, MMS-ViT-Base outperformed SATRN on IIIT, SVT, SP, and CT datasets by 3.6\%, 2.3\%, 2.6\%, and 3.9\%, respectively, with almost the same model structure and the number of parameters.
Compared with the state-of-the-art models, MMS-ViT-Base achieved competitive recognition accuracy with vision model SIGA (95.10\% vs. 95.58\%) and language model PARSeq (95.10\% vs. 95.28\%).
On the other hand, as described in \ref{sec:self-supervised text recognition}, when fine-tuned with ARD, the performance of MMS-Series significantly improved compared to fine-tuning with STD. MMS-ViT-Tiny outperforms the SOTA method SIGA by 1.73\%, and 2.16\% on Avg-IC13 and Avg-IC15, respectively, while MMS-ViT-Tiny has much fewer parameters than SIGA (20M vs. 113M).
Moreover, performance continued to improve as the model size increased, with MMS-ViT-Base surpassing SIGA by 2.26\% and 2.94\% on Avg-IC13 and Avg-IC15, respectively.

\subsubsection{Text Image Super-Resolution}
In Table~\ref{tab:sr result}, we evaluated the effectiveness of MMS pre-training in the text super-resolution task.
We compare MMS with self-supervised text recognition methods and previous state-of-the-art (SOTA) methods.
First, the comparison results of self-supervised methods are shown in the bottom section of Table~\ref{tab:sr result}. Our MMS showed significant improvement over Scratch and DiG in terms of SSIM and PSNR metrics.
When compared with the current SOTA method CCD, MMS  achieved a 0.59\% improvement in PSNR but a 0.45\% decrease in SSIM, resulting in competitive performance.
On the other hand, compared with previous SOTA super-resolution methods, our approach demonstrated superior performance in both PSNR and SSIM metrics. Notably, our text super-resolution model only employed three transformer units as the decoder following ViT-small, along with L2 loss. These experiments underscore the robust textual representation learning ability of our MMS in enhancing image quality.

\subsubsection{Text Segmentation}
In Table~\ref{tab:segmentation result}, we compare MMS with existing self-supervised text recognition methods in the text segmentation task. 
Compared to Scratch without pre-training, MMS exhibited a notable 6.9\% enhancement in IoU score.
Furthermore, MMS outperformed DiG by 1.9\% and marginally exceeded CCD by 0.2\%, which previously held the highest IoU score.
This experiment demonstrated MMS's capability to acquire features beneficial not only for text recognition but also for text segmentation.

\section{Conclusion}
\label{Conclusion}
In this study, we first analyzed different masking strategies for mask image modeling in textual representation learning. We found random masking predominantly learns low-level stroke (textural) information, while block and span masking learns relatively high-level character (contextual) information from unlabeled text images.
Taking into account the textural and contextual information inherent in text images, we proposed a novel self-supervised learning method for text recognition called Multi-Masking Strategy (MMS). 
MMS jointly utilizes multiple masking strategies to perform masked image modeling, enabling pre-trained models to learn semantic information that is useful for text recognition. 
Our comprehensive experimental results demonstrated that MMS outperforms the state-of-the-art self-supervised methods in various text-related tasks, including text recognition, text segmentation, and text image super-resolution when fine-tuned with real data.




\ifCLASSOPTIONcaptionsoff
  \newpage
\fi

\bibliography{bibtex/new_ref}
\bibliographystyle{IEEEtran}

\begin{IEEEbiography}[{\includegraphics[width=1in,height=1.25in,clip,keepaspectratio]{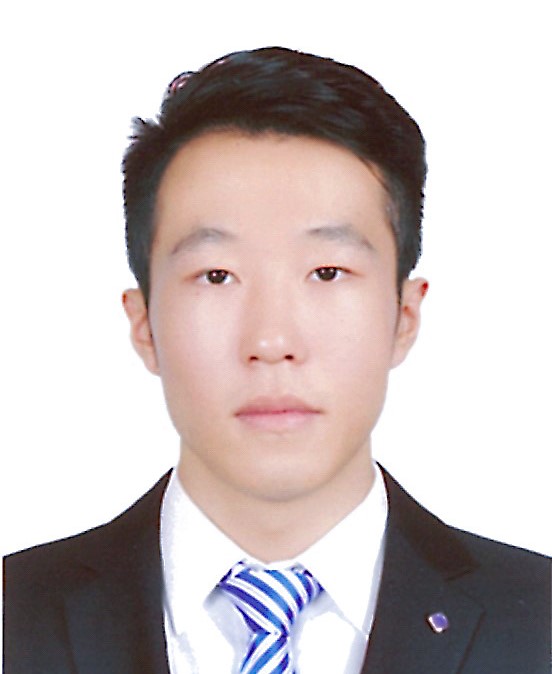}}]{Zhengmi Tang}
received his B.E., M.E., and Ph.D. degrees from Xidian University (2017), Hiroshima University (2020), and Tohoku University (2023), respectively. He is currently a researcher at the AIAMI of Wenzhou University, China. His current research interests include computer vision, scene text analysis, and data synthesis.
\end{IEEEbiography}

\begin{IEEEbiography}[{\includegraphics[width=1in,height=1.25in,clip,keepaspectratio]{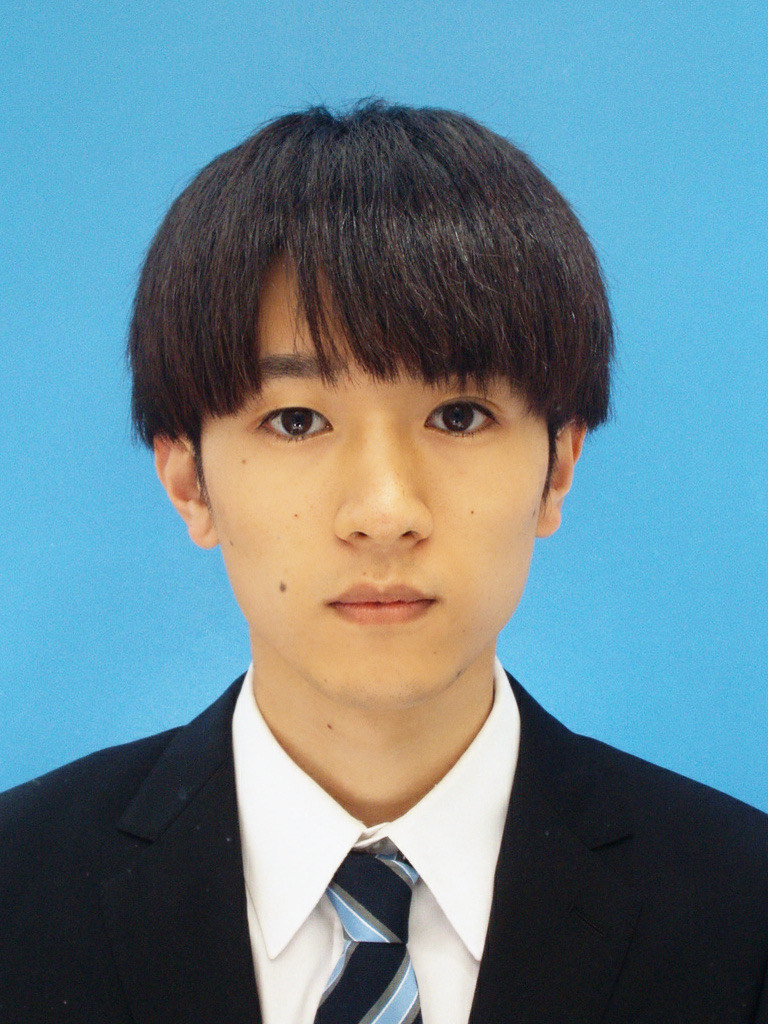}}]{Yuto Mitsui}
received the B.E. and M.E. degrees from Tohoku university, in 2022 and 2024, respectively. His research interest includes pattern recognition using deep neural networks.
\end{IEEEbiography}

\begin{IEEEbiography}[{\includegraphics[width=1in,height=1.25in,clip,keepaspectratio]{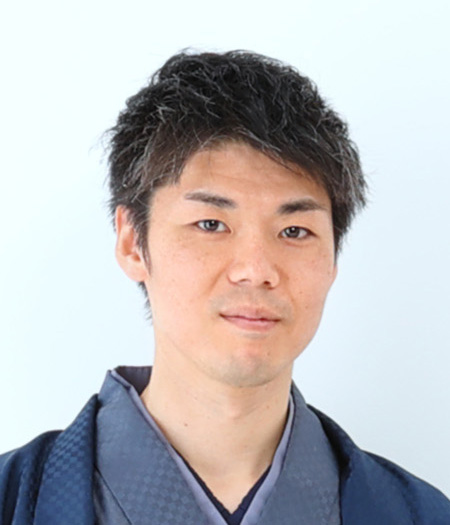}}]{Tomo Miyazaki}
 (Member, IEEE) received his B.E. and Ph.D. degrees from Yamagata University (2006) and Tohoku University (2011), respectively. From 2011 to 2012, he worked on the geographic information system at Hitachi, Ltd. From 2013 to 2014, he worked at Tohoku University as a postdoctoral researcher. Since 2015, he has been an Assistant Professor at the university. His research interests include pattern recognition and image processing.
\end{IEEEbiography}

\begin{IEEEbiography}[{\includegraphics[width=1in,height=1.25in,clip,keepaspectratio]{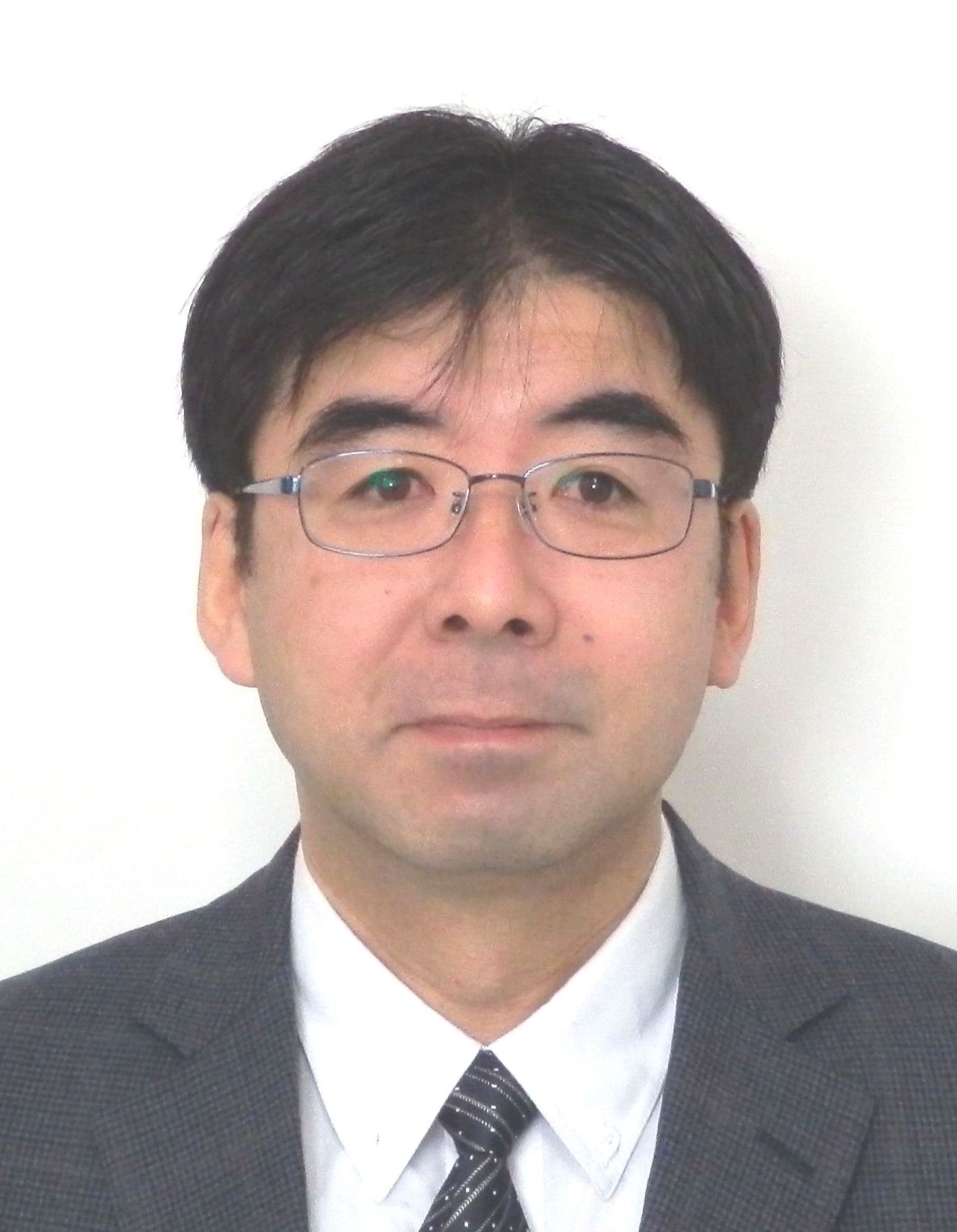}}]{Shinichiro Omachi}
(M’96-SM’11) received his B.E., M.E., and Ph.D. degrees in Information Engineering from Tohoku University, Japan, in 1988, 1990, and 1993, respectively. He worked as an Assistant Professor at the Education Center for Information Processing at Tohoku University from 1993 to 1996. Since 1996, he has been affiliated with the Graduate School of Engineering at Tohoku University, where he is currently a Professor. From 2000 to 2001, he was a visiting Associate Professor at Brown University. His research interests include pattern recognition, computer vision, image processing, image coding, and parallel processing. He served as the Editor-in-Chief of IEICE Transactions on Information and Systems from 2013 to 2015. Dr. Omachi is a member of the Institute of Electronics, Information and Communication Engineers, the Information Processing Society of Japan, among others. He received the IAPR/ICDAR Best Paper Award in 2007, the Best Paper Method Award of the 33rd Annual Conference of the GfKl in 2010, the ICFHR Best Paper Award in 2010, and the IEICE Best Paper Award in 2012. He served as the Vice Chair of the IEEE Sendai Section from 2020 to 2021.
\end{IEEEbiography}








\end{document}